\documentclass[review]{elsarticle}

\usepackage{graphicx}
\usepackage{amsmath}
\usepackage{amssymb}
\usepackage{times}
\usepackage{verbatim}
\usepackage{booktabs}

\begin{document}

\begin{frontmatter}



\title{Communities of Minima in Local Optima Networks of Combinatorial Spaces}


\author[label1]{Fabio Daolio}
\author[label1]{Marco Tomassini}
\author[label2]{S\'ebastien V\'erel}
\author[label3]{Gabriela Ochoa}

\address[label1]{Faculty of Business and Economics, Department of Information Systems, University of Lausanne, Switzerland}
\address[label2]{INRIA Lille - Nord Europe and University of Nice Sophia-Antipolis~/~CNRS, Nice, France}
\address[label3]{Automated Scheduling, Optimisation and Planning (ASAP) Group, School of Computer Science, University of Nottingham, Nottingham, UK}

\begin{abstract}

\noindent In this work we present a new methodology to study the structure of the configuration spaces of hard combinatorial
problems. It consists in building the network that has as nodes the locally optimal configurations and
as edges the weighted oriented transitions between their basins of attraction. 
We apply the approach to the detection of communities in the optima networks produced
by two different classes of instances of a hard combinatorial optimization problem: the quadratic
assignment problem (QAP).
We provide evidence indicating that the two problem instance classes give rise to very different configuration spaces.
For the so-called real-like class,  the networks possess
a clear modular structure, while the optima networks belonging to the class of random uniform instances are less well 
partitionable into clusters. This is convincingly supported by using several statistical tests.
Finally, we shortly discuss  the consequences of the findings for heuristically searching the corresponding problem spaces.

\end{abstract}

\end{frontmatter}


\section{Introduction}
\label{intro}

In the last decade some researchers have proposed a network view of energy landscapes in chemical physics
for small atomic clusters and macromolecules~\cite{scala,doye02,rao}. The idea is simple: if one can identify both the minima
of these energy landscapes and the possible transitions between them, then one can build a
graph in which the nodes are the minima and the transitions are represented by edges joining the corresponding minima~\cite{doye02}.
The number of minima usually increases exponentially with system size but significant samples can be obtained either by 
experiment or, much more often, by sampling the 
landscape using molecular dynamics and Monte Carlo methods (see~\cite{wales} and references therein). Doye has called this
graph the \textit{Inherent Structure Network}; here we prefer to use the term \textit{Local Optima Network} (LON).

Inspired by this approach, in our own work~\cite{pre09,ieee-neutral} we have applied similar ideas to the case of the configuration
spaces of difficult combinatorial optimization problems. Although the problems look superficially similar, hard combinatorial spaces pose
additional challenges due to their discrete nature, e.g. no derivatives and gradient information is available.
Moreover, the landscapes are rugged, may show frustration, i.e. not all local constraints can be satisfied together in order to
improve the objective function value,  and often contain neutrality which means that there are sizable regions where the objective function
values are identical or very close to each other. The typical physical models that show most of these features are spin 
glasses~\cite{spin-glass-book}
but the phenomenon is ubiquitous in combinatorial spaces of computationally hard problems~\cite{garey-johnson}.

Assuming that the LON of the given combinatorial landscape is known, or at 
least a significant sample has
been obtained, several questions are of great interest, both theoretical and practical. First and foremost, the distribution over the search space
of the optima and their connectivity, as well as the associated basins of attraction and their sizes, are fundamental information
that may help to understand the difficulty of the corresponding problem and may guide local search methods through
the problem configuration space. Second, the number and strength of possible transitions between basins, and
thus between optima, are also extremely important to gauge the stability of local optima configurations and the
possibility of jumping out of a suboptimal configuration to reach a better one, or even the globally optimal one.
Several other purely topological features of such a landscape representation are also very useful. For example, the
mean path length between any local optimum to the global one contains information about the computational
difficulty of the problem. The clustering coefficient of the optima network is also interesting as it gives information
about the topological transitivity between related optima.

In the present study we focus on a particular fundamental characteristic of the optima networks
using the \textit{Quadratic Assignement Problem} (QAP), which is a typical member of the class of computationally hard problems~\cite{garey-johnson}. The feature we are interested in, is related to the way in which
local optima are distributed in the configuration space. Then, several questions can be raised. Are they uniformly distributed as some
theoretical analyses of
fitness landscapes seem to assume for mathematical simplicity~\cite{garnier01}, or do they cluster in some non-homogeneous way?
If the latter, what is the relation between objective function values within and among different clusters and how easy is it to
go from one cluster to another? Knowing even approximate
answers to some of these questions would be very useful to further characterize the difficulty of a class of problems and also, potentially,
to devise new search heuristics or variation to known heuristics that take advantage of this information. 

There exist some measures that are used to characterize numerically distributions of landscape features
such as average solution distance and average distance between optima, where the distance is seldom the usual
Euclidean one but rather another kind of distance such as Hamming distance for problems defined on binary string
configurations, or the distance between two permutations in a permutation space~\cite{talbi-book}. However, a purely topological
vision may offer several advantages with respect to such ``metric'' approaches since in the former only information about the vertices
of the graph and their connections is needed. In complex network theory language, the above corresponds to the detection of 
\textit{communities} in the relevant LON networks. Community detection is a difficult task, but today several good
approximate algorithms are available and the approach is feasible, especially for the small networks
studied here~\cite{santo1}. A similar investigation has been performed
by Massen and Doye~\cite{massen-doye-comm} for the graphs of energy landscapes of small atomic clusters and by
Gfeller at al. for continuous free-energy landscapes in biomolecular studies~\cite{gfeller-pnas-07}.

The present study is structured as follows. In order to make the article sufficiently self-contained, we first briefly describe the QAP problem
and introduce our concepts  and methods to define and find the LONs.
 We next describe in detail the community search approach and we discuss the results obtained on two different classes of QAP problem instances and their significance. Finally, we give our conclusions.

\section{The QAP Problem}
\label{qap}

The QAP deals with the relative location of units that interact with one another in some manner. The objective is to minimize the total cost of interactions. The problem can be stated in this way: there are $n$ units or facilities to be assigned to $n$ predefined locations, where each location can accommodate any one unit. Location $i$ and location $j$ are separated by a distance $a_{ij}$, generically representing the per unit cost of interaction between the two locations. A flow of value $b_{ij}$ has to go from unit $i$ to unit $j$; the objective is to find and assignment, i.e. a bijection from the set of facilities onto the set of locations, which minimizes the sum of products flow $\times$ distance.

This bijection can naturally be encoded as a permutation $\pi$ and the problem of cost assignment  minimization can mathematically be formulated as:
\begin{equation}
\min_{\pi \in P(n)} C(\pi)=\sum_{i=1}^{n}\sum_{j=1}^{n}{a_{ij}b_{\pi_{i}\pi_{j}}}
\label{eq:fitness}
\end{equation}

\noindent where $C(\pi)$ is the cost function, $A=\{a_{ij}\}$ and $B=\{b_{ij}\}$ are the two $n \times n$ distance and flow matrixes, $\pi_{i}$ gives the location of facility $i$ in permutation $\pi \in \Sigma_n$, and $\Sigma_n$ is the set of all permutations of $\{1,2,...,n\}$, i.e. the QAP search space.
The structures of the distance and flow matrices characterize the class of instances of the QAP problem. Later in the article it is explained which are the classes of instances used in the present work.

\section{Local Optima Networks}
\label{LON}

Given a fitness landscape for an instance of a combinatorial optimization problem like the QAP, we have to define the associated optima network by providing definitions for the nodes and the edges of the network. The vertexes of the graph can be straightforwardly defined as the local minima of the landscape. In this  work,  we select small QAP instances such that it is feasible to obtain the nodes of the graph exhaustively by running a best-improvement local search algorithm from every configuration of the search space as described below. Before explaining how the edges of the network are obtained,  a number of relevant definitions are summarized.

A Fitness landscape~\cite{stadler-02} is a triplet $(S, V, f)$ where $S$ is a set of potential solutions, i.e. a search space, $V : S \longrightarrow 2^S$, a neighborhood structure, is a function that assigns to every $s \in S$ a set of neighbors $V(s)$, and $f : S \longrightarrow \mathbb{R}$ is a fitness function, also called cost function or objective function, that can be pictured as the \textit{height} of the
corresponding solutions. In our study, a search space configuration $s$  is a permutation $\pi$ of the $n$ facility locations, therefore the search space size is $n!$.  The neighborhood of a configuration is defined
by the pairwise exchange operation, which is the most basic operation used by many meta-heuristics for QAP. This operator simply exchanges any two positions in a permutation $\pi$, thus transforming it  into another permutation. The neighborhood size is thus $|V(s)| = n(n-1)/2$. 
Finally, the fitness for this problem is defined by equation~\ref{eq:fitness} as $f(s) = -C(s)$.

The \textit{Best Improvement} (BI) algorithm to determine the local optima and
therefore define the basins of attraction starts from an arbitrary configuration $s$ and systematically tries to improve
the solution by looking at all neighbor solutions $V(s)$, choosing the best one. It stops when no improvement is possible, i.e. when the current solution
$s^{*}$ is a local optimum: $\forall  s \in
V(s^{*})$, $f(s) < f(s^{*})$.

The basin of attraction of a local optimum $i \in S$ is the 
set $b_i = \{s \in S ~|~ BI(s) = i \}$. The size of the basin of
attraction of a local optimum $i$ is the cardinality of $b_i$.
The basins of attraction as defined above produce a
partition of the configuration space $S$. Therefore, $S = b_1 \cup b_2 \cup \ldots \cup b_n$
 and $\forall i \not= j$, $b_i \cap
b_j = \emptyset$.\\
We can now define the weight of an edge that connects two feasible solutions in the
 fitness landscape.
\noindent For each pair of solutions $s$ and $s^{'}$, $p(s
\rightarrow s^{'} )$ is the probability
to go from $s$ to $s^{'}$ with the given neighborhood structure.
For the search space of permutations of $n$ elements, and the pairwise exchange operation, there are $n(n-1)/2$ neighbors for each solution, therefore:

\noindent if $s^{'} \in V(s)$ , $p(s \rightarrow s^{'} ) = \frac{1}{n(n-1)/2}$ and \\
if $s^{'} \not\in V(s)$ , $p(s \rightarrow s^{'} ) = 0$.

\noindent The probability of going from a solution $s
\in S$ to a solution belonging to the basin $b_j$, is defined as:
$$
p(s \rightarrow b_j ) = \sum_{s^{'} \in b_j} p(s \rightarrow s^{'} ).
$$

\noindent Notice that $p(s \rightarrow b_j ) \leq 1$.
Thus, the total probability of going from basin $b_i$ to
basin $b_j$ is the average over all $s \in b_i$ of the transition
probabilities  to solutions $s^{'} \in b_j$ :

$$p(b_i \rightarrow b_j) = \frac{1}{|b_i|} \sum_{s \in b_i} p(s \rightarrow b_j ),$$

\noindent where $|b_i|$ is the size of the basin $b_i$.

\noindent Now we can define a \textit{Local Optima Network} (LON)
as being the graph  $G=(S^*,E)$ where the set of vertices $S^*$
contains all the local optima, and there is an edge $e_{ij} \in E$ with
weight $w_{ij} = p(b_i \rightarrow b_j)$ between two nodes $i$ and
$j$ iff $p(b_i \rightarrow b_j) > 0$.
Notice that since each maximum has its associated basin, $G$ also
describes the interconnection of basins.

\noindent According to our definition of edge weights, $w_{ij} = p(b_i
\rightarrow b_j)$ may be different than $w_{ji} = p(b_j \rightarrow
b_i)$. Thus, two weights are needed in general, and we have an
oriented transition graph.

\section{Structure of the QAP LONs}

\subsection{Problem Instance Generation}
In order to perform a statistical analysis,  several problem instances of at least two different problem classes have to be considered. To this purpose, the two instance generators proposed by Knowles and Corne~\cite{Knowles2003emo} for the multi-objective QAP have been adapted and used here for the single-objective QAP.
The first generator produces uniformly random instances where all flows and distances are integers sampled from uniform distributions.
 This leads to the kind of problem known in literature as \emph{Tai}\verb"nn"\emph{a}, being \verb"nn" the problem dimension~\cite{Taillard1995}. Distance matrix entries are, in both cases, the Euclidean distances between points in the plane.
The second generator permits to obtain  flow entries that are non-uniform random values. 
This procedure, detailed in~\cite{Knowles2003emo} produces random instances of type \emph{Tai}\verb"nn"\emph{b} which have the so called ``real-like'' structure since they resemble to the structure of QAP problems found in practical applications.
For a general network analysis, 30 random uniform and 30 random real-like instances have been generated for each problem dimension in $\{5,...,10\}$.
To the specific purpose of community detection, 200 additional instances have been produced and analyzed with size $9$ for the random 
uniform class, and size $11$ for the real-like instances class. 
Problem size $11$ is the largest one for which an exhaustive sample of the configuration space is computationally feasible.
Beyond that, sampling must be used. However, in this work we prefer to stick with exact results in order to give as accurate as
possible answers
to the minima clustering problem posed at the beginning.

\subsection{Network Analysis}
\label{net-analys}

The results of the statistical analysis of the above QAP landscapes, up to size $10$, appear in~\cite{QAP-cec-10} to which the reader is referred for further information. 
In that work it is shown that LONs for the QAP are dense, as one can see from Tables~\ref{Rout/Nv} and~\ref{Rout/Ne} which give,
respectively, the mean number of vertices and the mean number of edges for the two classes of instances and for instance
sizes going from $5$ to $10$.
\begin{table}[hptb]
\begin{center}
\caption{Average values of the number of vertices for each instance size.\label{Rout/Nv}} 
\vspace{3mm}
\begin{tabular}{ccccccccc} \toprule
\multicolumn{1}{}{}&\multicolumn{1}{c}{5}&\multicolumn{1}{c}{6}&\multicolumn{1}{c}{7}&\multicolumn{1}{c}{8}&\multicolumn{1}{c}{9}&\multicolumn{1}{c}{10}\\ \cmidrule{2-7}
real-like&~~1.667 &~~2.767&~~3.900&~~6.133 &~12.567 &~25.700\\ 
uniform&~~3.333&~~6.800&~19.100&~51.300&137.300&414.133\\ 
\bottomrule
\end{tabular}
\end{center}
\end{table}

\begin{table}[hptb]
\begin{center}
\caption{Average value of the number of edges for each instance size.\label{Rout/Ne}} 
\vspace{3mm}
\begin{tabular}{ccccccccc} \toprule
\multicolumn{1}{}{}&\multicolumn{1}{c}{5}&\multicolumn{1}{c}{6}&\multicolumn{1}{c}{7}&\multicolumn{1}{c}{8}&\multicolumn{1}{c}{9}&\multicolumn{1}{c}{10}\\ \cmidrule{2-7}
real-like & 3.400 & 9.433 & 19.900 & 46.0667 &187.433 & 818.700\\ 
uniform & 12.600 & 50.733 & 399.767 & 2798.233 &19225.370 & 169118.800\\ 
\bottomrule
\end{tabular}
\end{center}
\end{table}

Table~\ref{Rout/EtoV2} shows that the LONs of both the uniform and real-like instances up to size $10$ are complete
or almost complete since $|E| = O(|V|^2)$. This is inconvenient for community analysis as it is difficult for any cluster detection algorithm
to split-up the networks into separate communities when the graphs are so dense.

\begin{table}[hptb]
\begin{center}
\caption{Average of the ratio of the number of edges to squared number of vertices\label{Rout/EtoV2}} 
\vspace{3mm}
\begin{tabular}{ccccccccc} \toprule
\multicolumn{1}{}{}&\multicolumn{1}{c}{5}&\multicolumn{1}{c}{6}&\multicolumn{1}{c}{7}&\multicolumn{1}{c}{8}&\multicolumn{1}{c}{9}&\multicolumn{1}{c}{10}\\ \cmidrule{2-7}
real-like &1.000 &0.993&0.994&0.999&0.992&0.988\\ 
uniform &0.998&0.993&0.969&0.940&0.9087&0.874\\ 
\bottomrule
\end{tabular}

\end{center}
\end{table}

\begin{figure}[h!]
\begin{center}
 \includegraphics[width=0.49\textwidth]{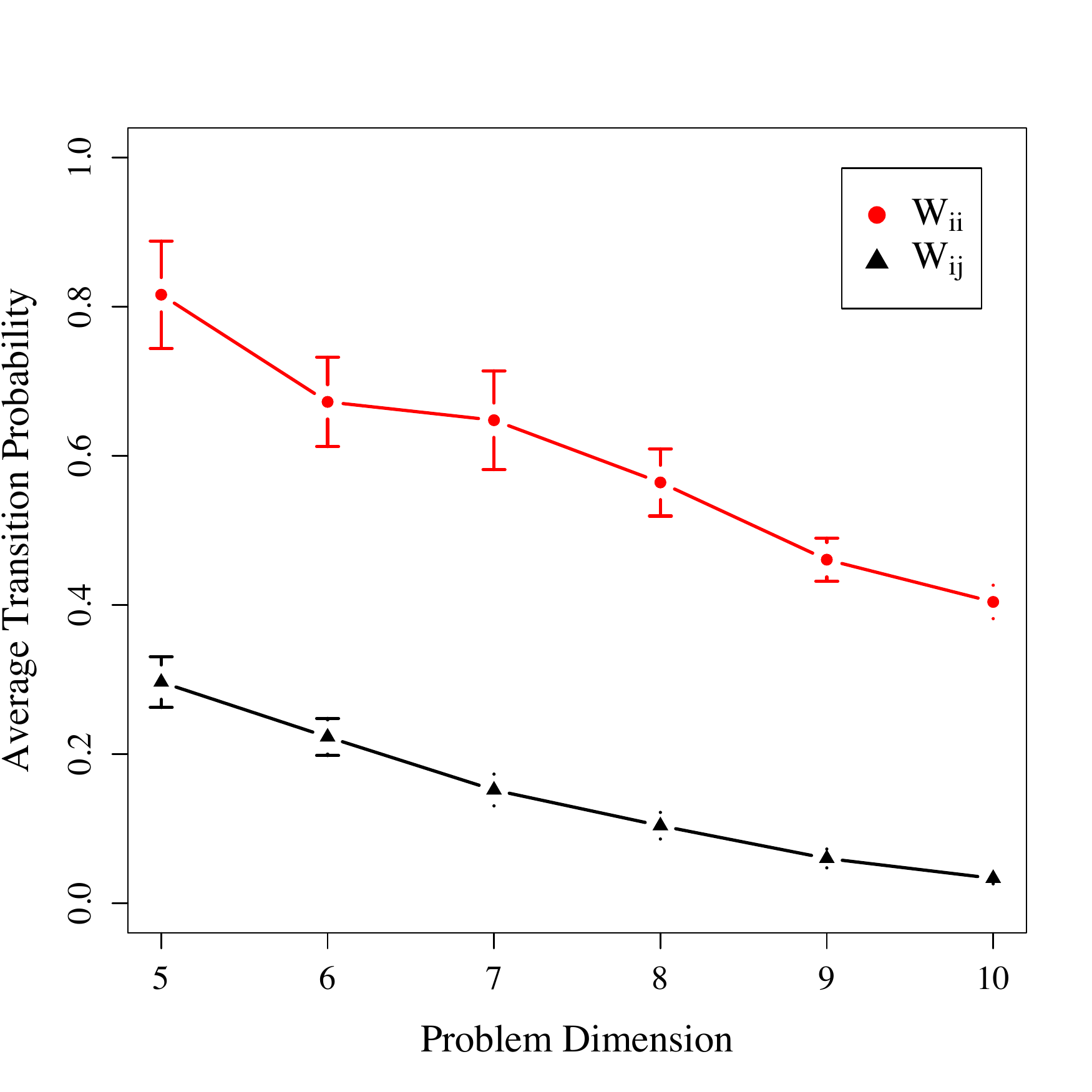}
 \includegraphics[width=0.49\textwidth]{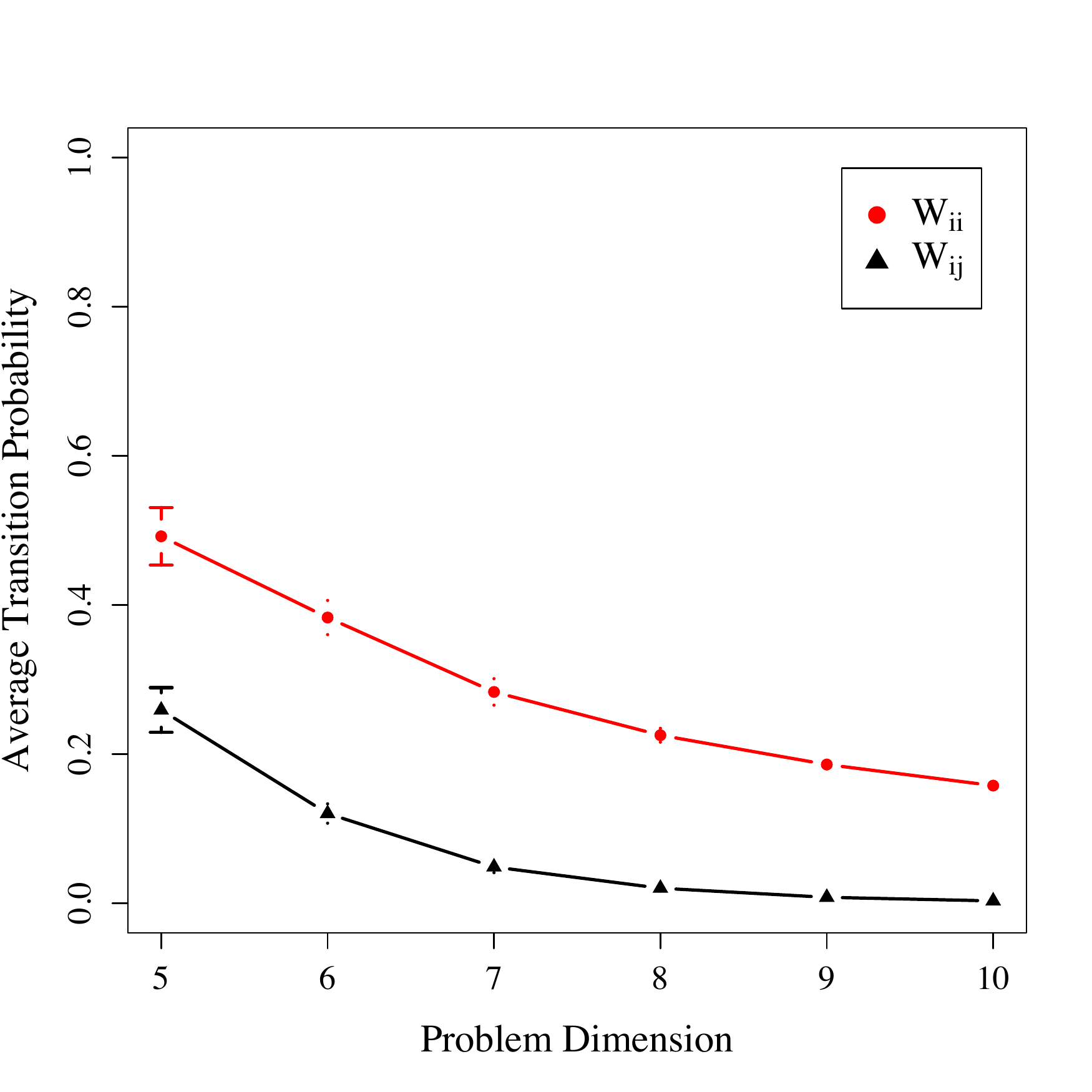}
  \vspace{-0.3cm}
 \caption{Average weights $w_{ii}$ for self-loops (circular points) and $w_{ij}$ for out-going links (triangular points). Left image: real-like instances. Right image: uniform instances. 
 Bars depict $95\%$ confidence intervals on the means (standard errors).
 Averages from 30 independent and randomly generated instances are shown.}
\label{fig:wii}
\end{center}
\end{figure}

However, looking at Fig.~\ref{fig:wii} which gives the average values of the transition probabilities to stay in the
same basin and to jump to another basin,  it is apparent that most of the probability distribution
is placed in $p(i \rightarrow i )$, i.e. it is more likely for a solution to stay in the same basin rather than to jump to
a neighboring one for both instance classes. When a local stochastic heuristic is used to search the landscape, 
only transitions to
another basin that have higher probability of occurring are important. Actually, a search heuristic like simulated annealing
also accepts moves that worsen the objective function with low probability, but the acceptance rate decays quickly as
the temperature is lowered and, in the end, only the most likely transitions play a role. In fact, the important role that
``weak ties'' may have in a social network context~\cite{granovetter} is almost absent in combinatorial landscapes where the more probable
search paths are associated with the highest transition probabilities when these landscapes are searched with a local
stochastic heuristic.
These considerations give us a clue as to
how to filter out the network edges in such a way that only the more likely transitions are kept and, as a consequence,
the graph becomes much less dense and gives a coarser but clearer view of the fitness landscape backbone. Such a network can be 
used for community analysis. On the other hand, it is also possible to keep the original dense weighted directed networks
and to use a specialized community detection algorithm that was originally designed to work in such cases, which
is called the \textit{Markov Clustering Algorithm} (MCL)~\cite{MCL}. We first explain our filtering procedure in the following
and then we will compare the results with those obtained through the MCL algorithm.
Nevertheless, we point out that, even in the MCL case, the algorithm itself does some preprocessing of the network weights.

The filtering procedure is very simple. First, we transform the weighted directed graph $G$ into a weighted undirected one $G_u$ by 
taking each pair of edges $\vec{ ij}$ and $\vec{ji}$ and replacing them with a single undirected link $ij$ whose weight is
the average:

$$ w_{ij} = \frac {\vec{w}_{ij} + \vec{w}_{ji}} {2}. $$

\noindent Indeed, the values of $\vec{w}_{ij}$ and $\vec{w}_{ji}$ are different in general and thus taking a single edge is an
approximation. However, a local heuristic walking the landscape would still be able to traverse a link in both directions, whereas
filtering the directed edges could cause some transitions to disappear.

Now, on $G_u$ we establish a probability threshold $\Pi$ for the weights $w_{ij}$ associated
to each edge  in $G_u=(S^*,E)$ and suppress all edges  that have $w_{ij}$ smaller than the value
marking the $\Pi$-quantile in the weights distribution.
We call the resulting
network $G_{u}^{'} = (S^*,E^{'})$; it has the same number of vertices as $G_u$ and a number of edges $|E^{'}| \le |E|$.
 
The following Figures~\ref{figpi1} and~\ref{figpi2} show an example of the results of such a filtering process, starting from the LON of a particular instance and for four values of the threshold $\Pi$.

\begin{figure}[h!]
\begin{center}
 \includegraphics[width=0.47\textwidth]{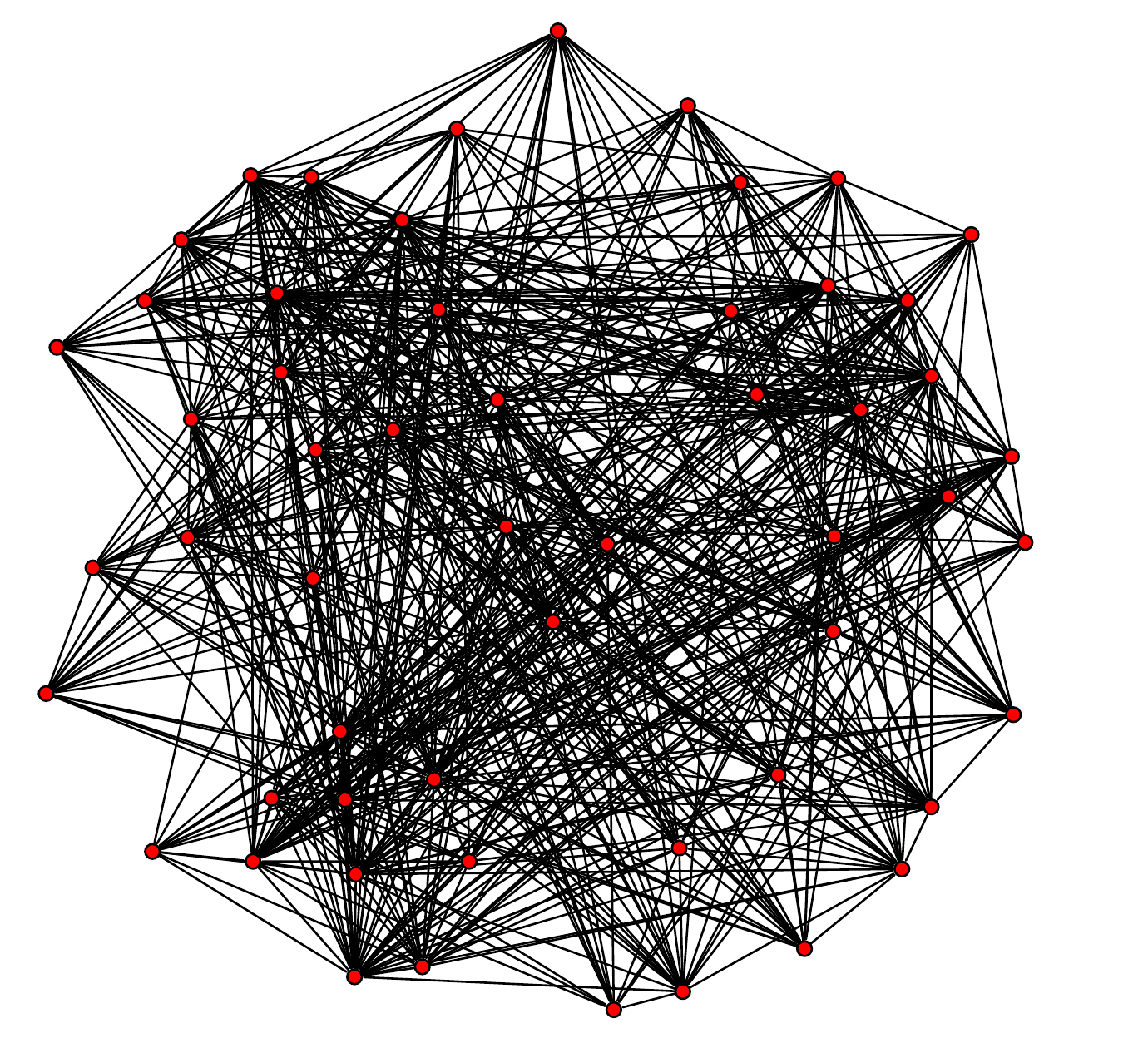}
 \includegraphics[width=0.47\textwidth]{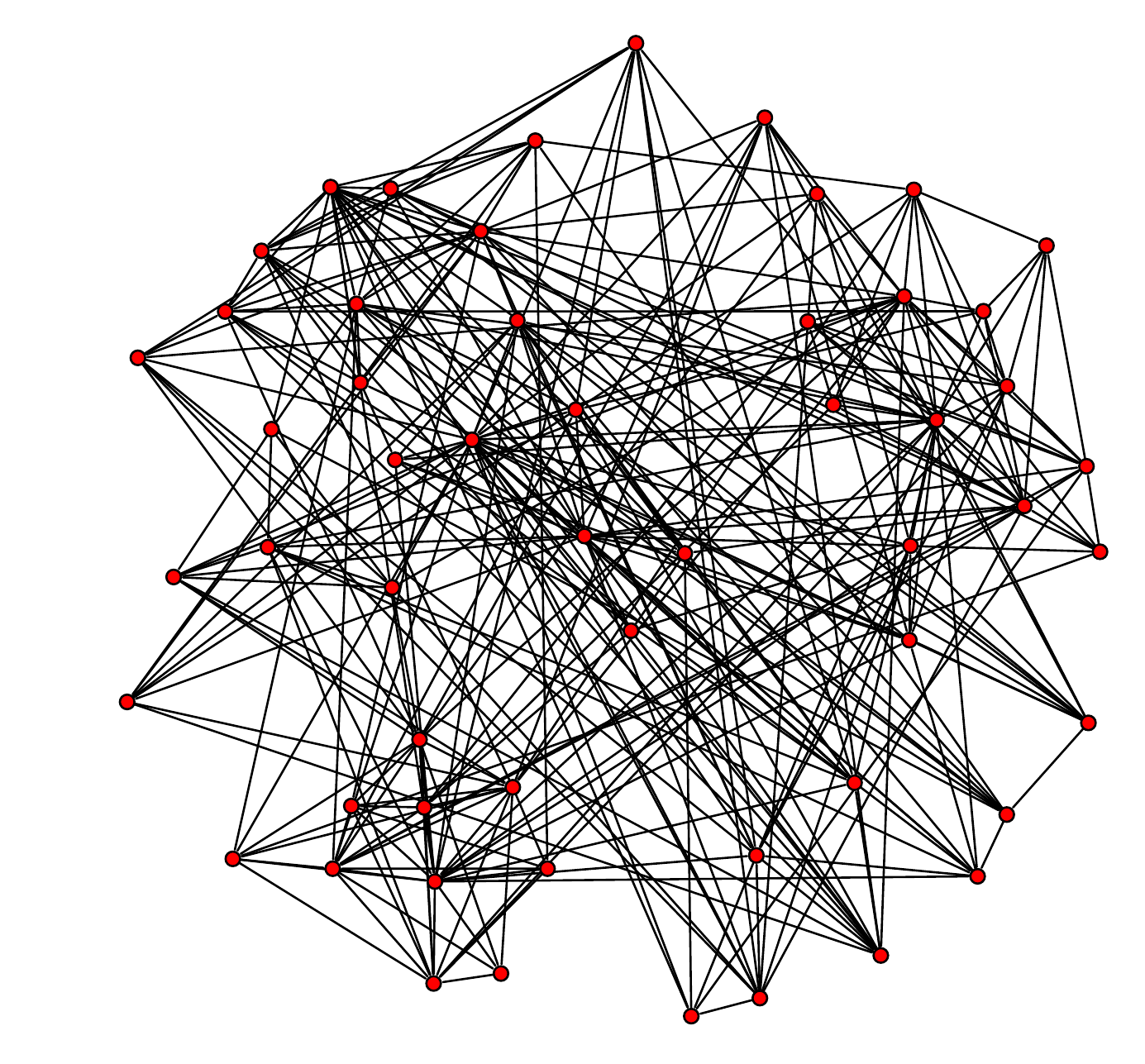}
  \vspace{-0.3cm}
 \caption{Left image, edge pruning threshold $\Pi=0.50$, right image $\Pi=0.75$. }
\label{figpi1}
\end{center}
\end{figure}

\begin{figure}[h!]
\begin{center}
 \includegraphics[width=0.48\textwidth]{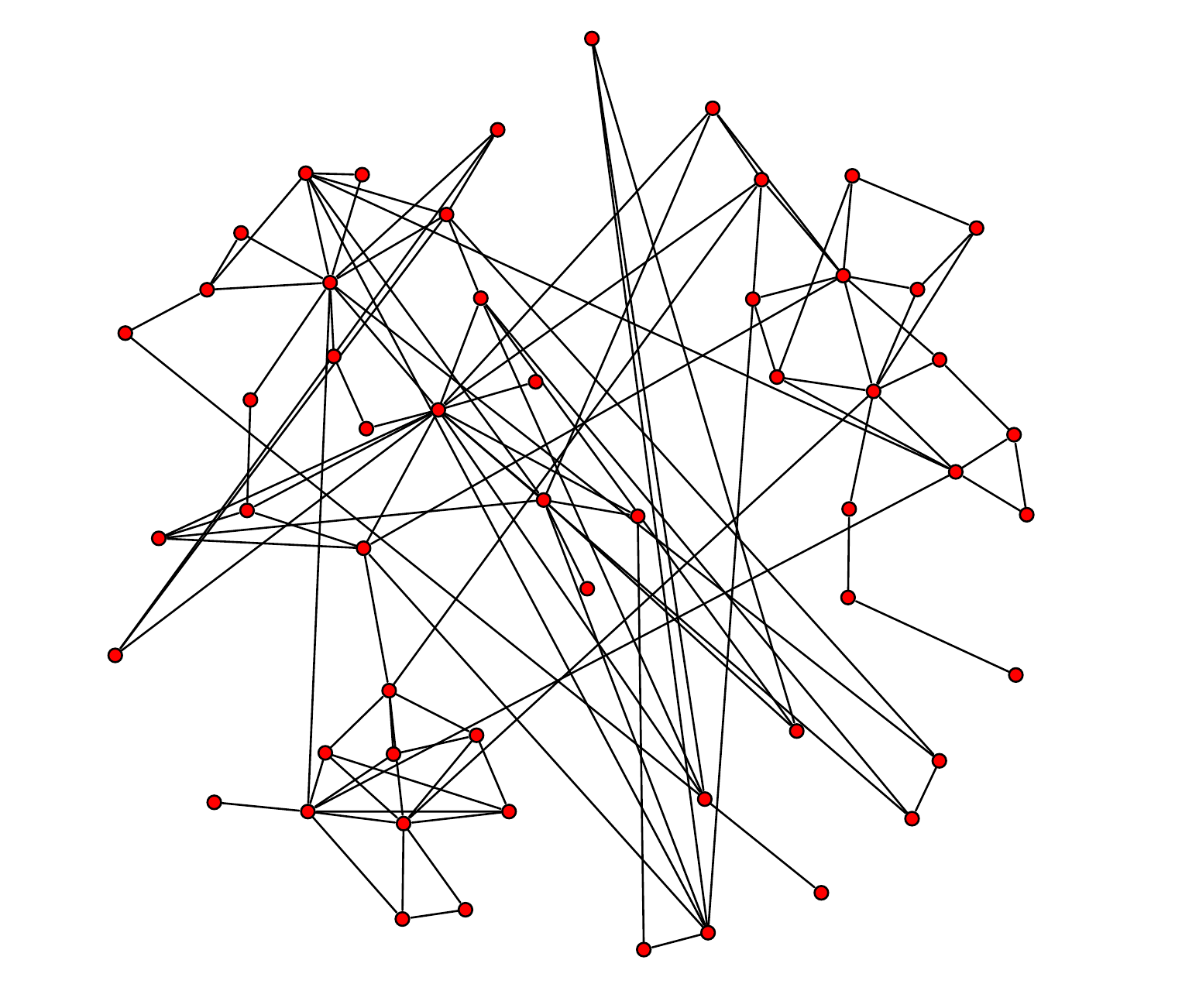}
 \includegraphics[width=0.45\textwidth]{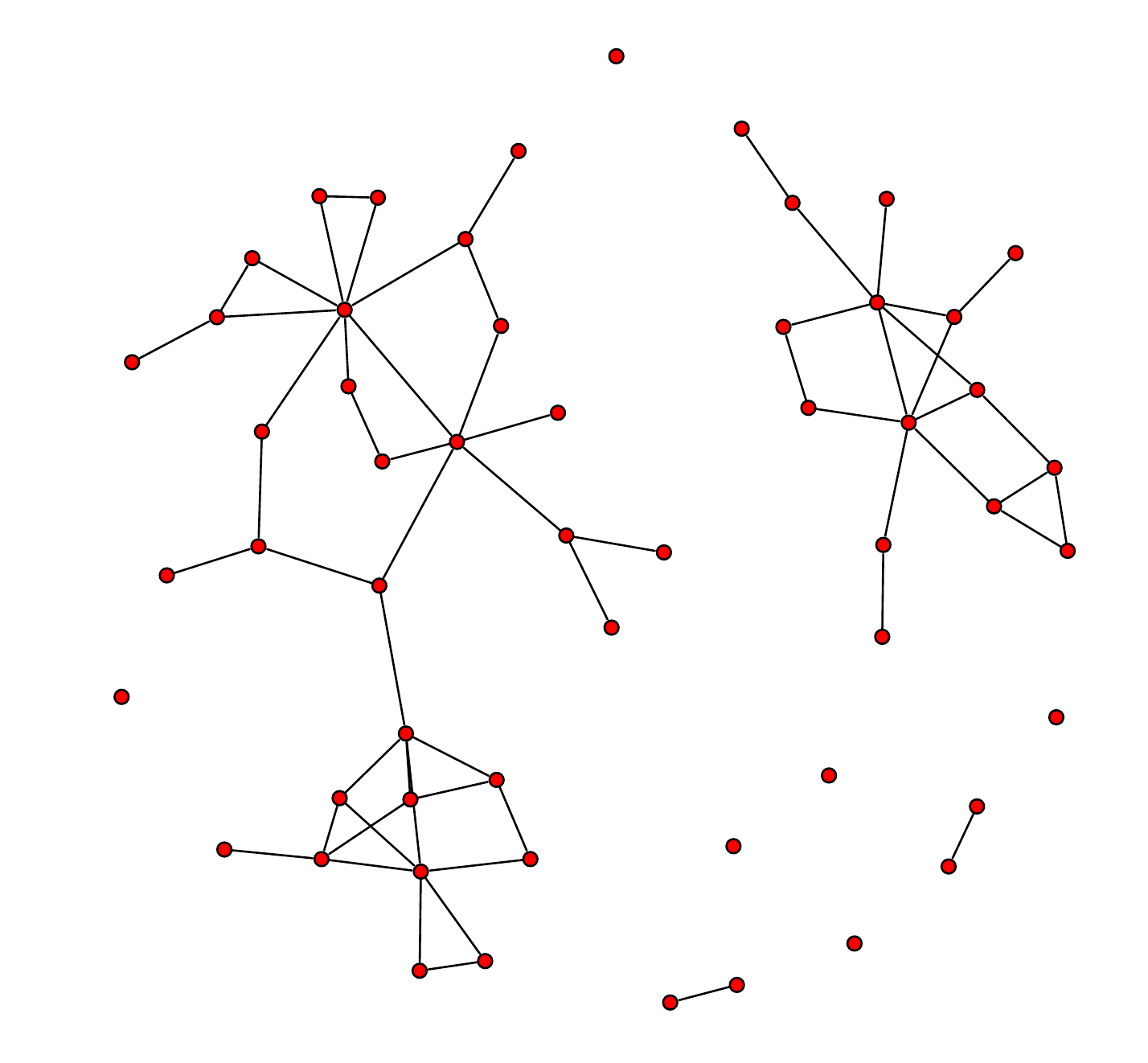}
  \vspace{-0.3cm}
 \caption{Left image, Threshold $\Pi=0.91$, right image $\Pi=0.95$; in this case the graph becomes disconnected. }
\label{figpi2}
\end{center}
\end{figure}

How to choose the threshold $\Pi$ is essentially a matter of trial and error. Too low a value does not allow a crisp picture
of the network to emerge; but if $\Pi$ is too high the network finally falls apart into separated components as in Fig.~\ref{figpi2}
 which is not
acceptable since, by definition of fitness landscape, the whole graph must be connected. Likewise, if some node becomes
isolated during filtering, it cannot be simply removed since it could, at least in principle, be the configuration corresponding to
the global optimum. Thus, each analyzed network has been filtered up to the maximum value of $\Pi$ that still
preserves its connectivity.

\subsection{Communities of Optima in LONs}

Communities or clusters in networks can be loosely defined as being groups of nodes that are
strongly  connected between them and poorly connected with the rest of the graph. 
Several methods have been proposed to uncover the clusters present in a network
(for an excellent recent review see~\cite{santo1}). Community detection belongs to the class of graph
partitioning problems, which are hard in the sense that there is no known algorithm bounded by a
polynomial function of the size of the input to exactly solve the problem~\cite{garey-johnson}. Community detection has
the added difficulty that there is not a single accepted rigorous measure of a cluster or a partition of the nodes of
a given graph into meaningful clusters. One commonly used measure is \textit{Modularity}.
The \textit{modularity} $Q$ of a partition has been defined as a merit function measuring the fraction of within community edges minus the
expected value of such fraction for a randomized network with the same vertex degree distribution~\cite{PhysRevE.69.026113}.

Several heuristics have been proposed~\cite{santo1};
after a preliminary analysis, we have chosen two of them: Clauset et al's. method based on greedy modularity optimization~\cite{clauset2004finding},
and Reichardt's and Bornholdt's spin glass ground state-based 
algorithm~\cite{reichardt2006statistical}\footnote{For the actual computations and data treatment, 
the ``igraph'' complex network analysis package~\cite{csardi2006igraph}, along with
the R statistical environment~\cite{ihaka1996r}, has been used.}. Both methods gave consistent results on our networks and,
in addition, they also work with undirected weighted networks which was required in our case.
The reason for using two methods is that we can assess statistical significance independent of the algorithm
and we can double check the community partition results. 

\begin{figure}[h!]
\begin{center}
 \includegraphics[width=0.7\textwidth]{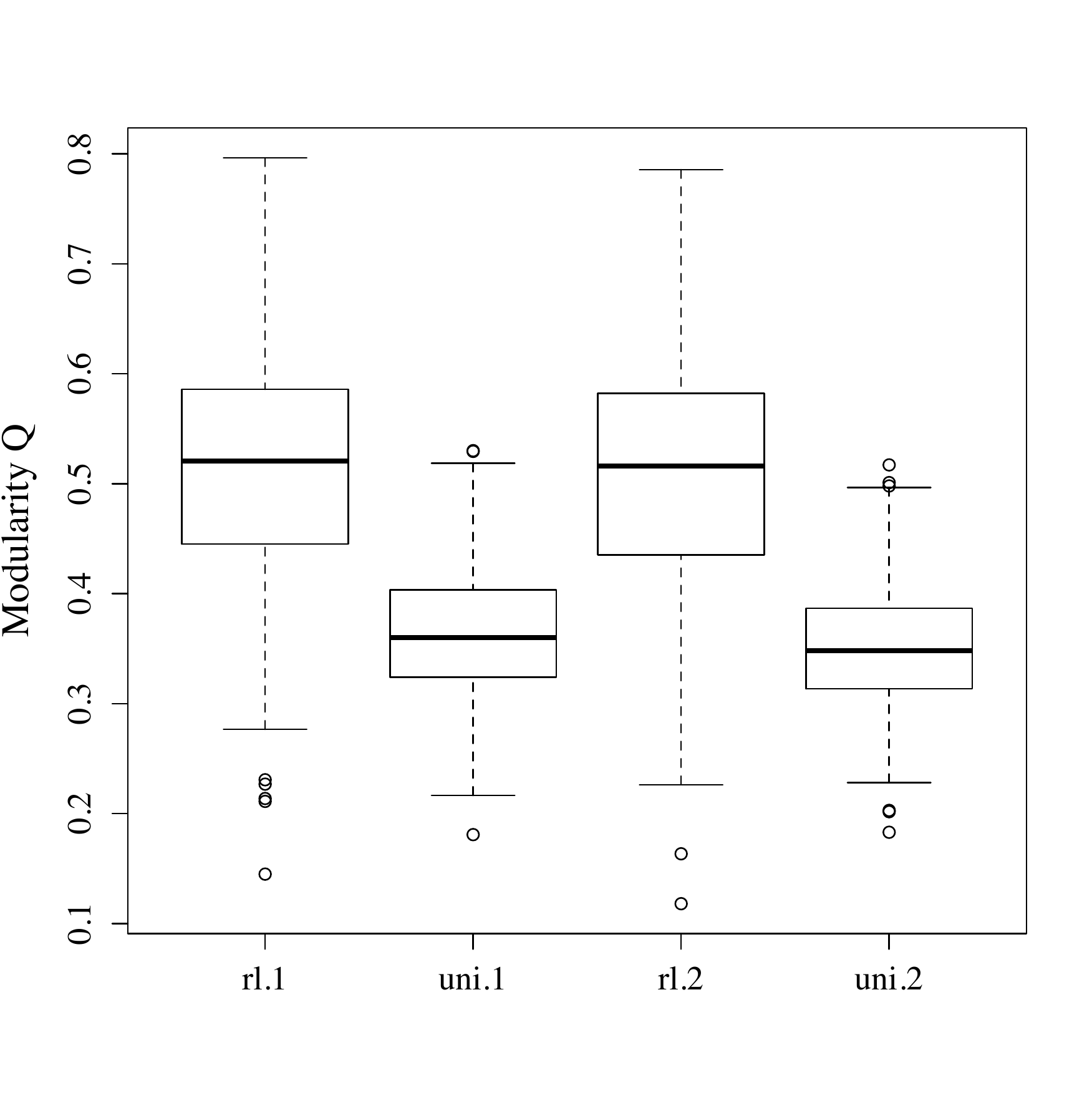}
  \vspace{-0.3cm}
 \caption{Boxplots of the modularity score $Q$ on the y-axis with respect to class problem (rl stands for real-like and
  uni stands for random uniform) and community detection algorithm (1 stands for fast greedy modularity
  optimization and 2 stands for spin glass search algorithm). 
  }
\label{box}
\end{center}
\end{figure}

For the sake of the community analysis we shall use only the results on newly generated
problems of size $11$
for the real-like instances and of size $9$ for the uniform random ones. This is suggested by the consideration
that the LONs for these two cases have comparable sizes in terms of number of vertices and can still be 
obtained exactly by an exhaustive search. 
The resulting
networks are still relatively small: the average size is $60$ for real-like instances and $127$ for random uniform ones but 
already sufficient for a meaningful cluster analysis. For example, they
have sizes larger than the famous ``Zachary's Karate Club Network''~\cite{karate}, which has $34$ nodes and
 is routinely mentioned as a standard benchmark in community detection work~\cite{santo1}.

Before examining the actual community structures found, we present the results of some tests in order to evaluate
the statistical significance of the clustering in terms of modularity. 

In general, the higher the value of $Q$ of a partition, the crisper the community structure.
Figure~\ref{box} is a plot of the modularity score $Q$ distribution empirically determined from the data for each algorithm/problem
class pair. The boxes ``hinges'' represent the $25$, $50$ (thick lines), and $75\%$ quantiles. The plot indicates that the
two problem classes are well separated in terms of $Q$, and that the community detection algorithm does not seem to
have any influence on such a result. To further show that these results are statistically significant, we have performed
a permutation test~\cite{manly2007randomization} for a factorial ANOVA design,
modeling the modularity scores as a response variable to the problem
class and algorithm choice. The $p$-values thus obtained are $2 \times 10^{-4}$, $0.179$, and $0.6414$ for
the factor ``problem class'', factor ``algorithm'', and the interaction between the two, respectively. Only the first one is below the significativity
threshold of $0.05$.
It is thus safe to conclude that the data show no significant effect of the community detection method used on the variability observed in modularity score, while the problem class only seems to explain that variability: real-like instances can be clustered in more modular partitionings than random uniform ones.

\begin{figure}[h!]
\begin{center}
 \includegraphics[width=0.8\textwidth]{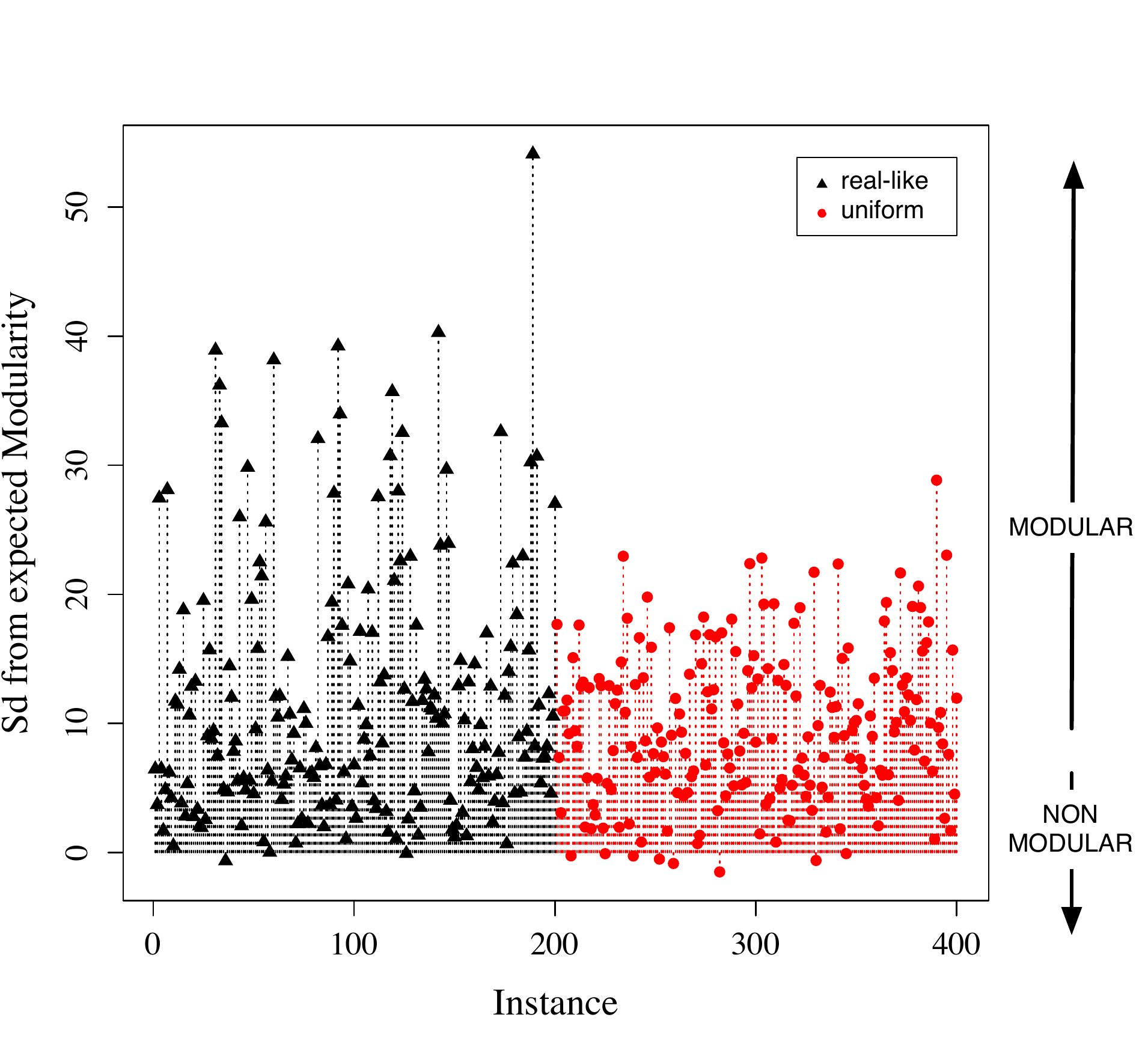}
  \vspace{-0.3cm}
 \caption{Modularity $Q$ for the best clustering into communities by means of greedy optimization algorithm. For each instance (x-axis: $200$ real-like ones on the left, $200$ random uniform ones on the right), $Q$ is plotted in number of standard deviations (y-axis) away from the average of $1000$ randomised networks with the same degree sequence.}
\label{devs1}
\end{center}
\end{figure}

\begin{figure}[h!]
\begin{center}
 \includegraphics[width=0.8\textwidth]{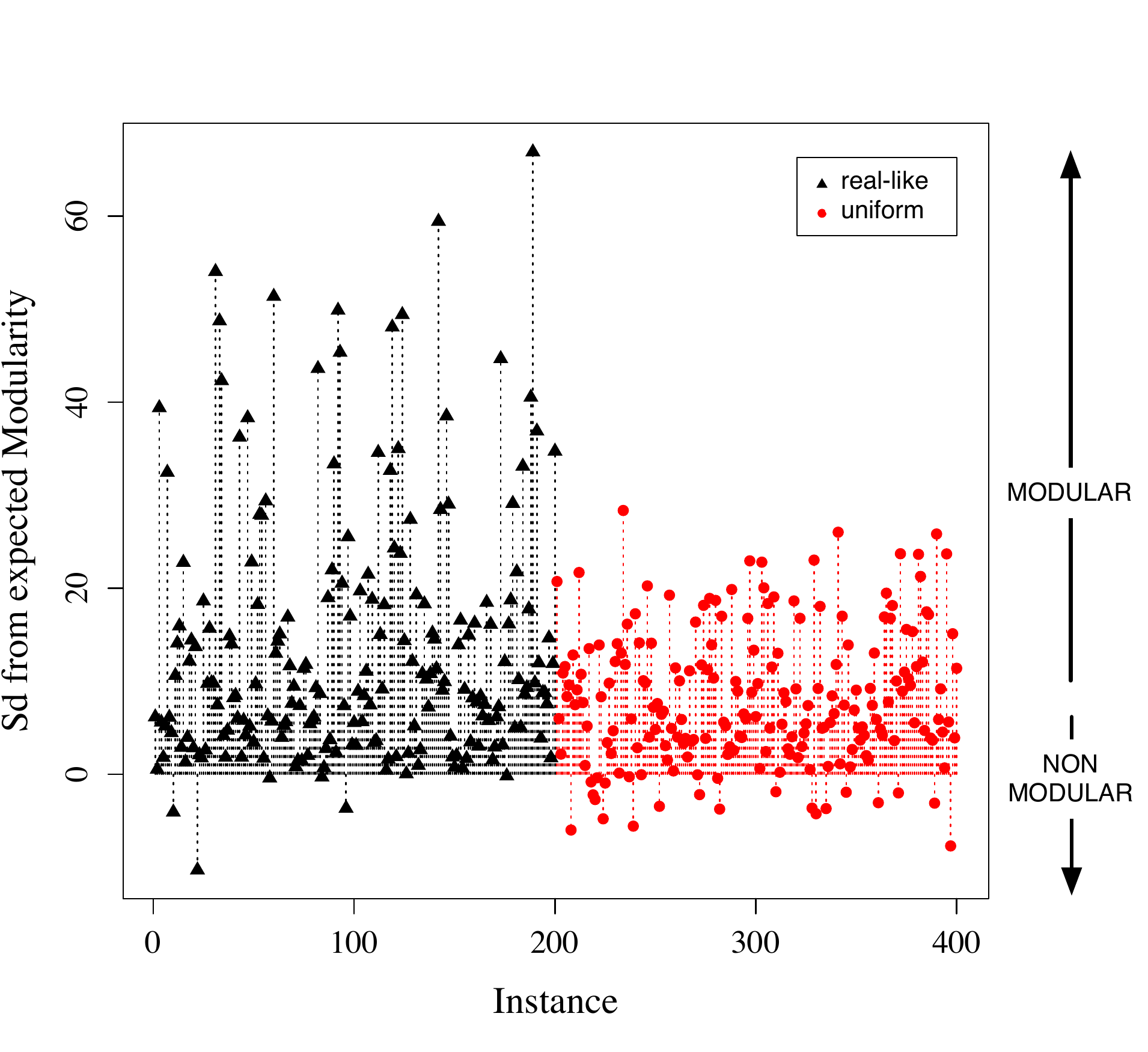}
  \vspace{-0.3cm}
 \caption{Modularity $Q$ for the best clustering into communities by means of spin glass model algorithm.
For each instance (x-axis: $200$ real-like ones on the left, $200$ random uniform ones on the right), $Q$ is plotted in number of standard deviations (y-axis) away from the average of $1000$ randomised networks with the same degree sequence.}
\label{devs2}
\end{center}
\end{figure}

However, modularity scores alone can be misleading. This has been shown by Guimer\`a et al. in~\cite{PhysRevE.70.025101} where
they pointed out that many random networks with no clear community structure may nevertheless have rather high
values of $Q$ due to statistical fluctuations. Thus, to test for the statistical significance of $Q$ we used a Monte Carlo
procedure in which a randomized version of the data is produced. For each problem instance, $1000$ random networks have
been generated using Viger's and Latapy's algorithm~\cite{viger2005efficient}, i.e. starting with the original graph's degree sequence and rewiring the links
randomly without altering the sequence. Next, both community detection algorithms have been applied to each generated
network to obtain the modularities $Q$ of their partitions. Finally, a $p$-value has been computed by comparing the expected
value of $Q$ from the randomized networks with the $Q$ measured for the actual original network.
Among the $200$ considered real-like instances, those $p$-values are not significant in $14$ and $22$ cases when using the two community finding algorithms. Among the $200$ random uniform instances, the non-significant cases are $16$ and $36$, respectively.
These figures are slightly higher than ``5-out-of-100" ratio one could expect from the significance threshold, but 
not by a margin high enough to invalidate the results of the whole analysis.
In conclusion, a separation between the two problem classes, with respect to minima clustering in the search space, has been significatively highlighted.


Once the significance of a community structure has been assessed, in order to have an insight into its strength, one can also measure for each instance how distant the obtained modularity score is from the expected value estimated on the respective null model.
Networks observed in nature and society present a modularity structure not only significantly different but markedly higher than random networks~\cite{guimera2006classes}.
In this respect, Figures~\ref{devs1} and~\ref{devs2} depict  for each instance the difference in number of standard deviations (with respect to the
$0$ horizontal line)
between the measured modularity scores and the expected ones for the null model (a population of $1000$ randomized networks having the same degree sequence, as explained above).
It is worth stressing that the absolute value of such a distance depends on the variability observed within the null models. Nonetheless,
there is  difference between the two problem classes, with real-like ones displaying a stronger and more instance-dependent community structure than random uniform ones, disregarding the algorithm chosen to discover such a structure.\\

\begin{figure}[h!]
\begin{center}
 \includegraphics[width=0.7\textwidth]{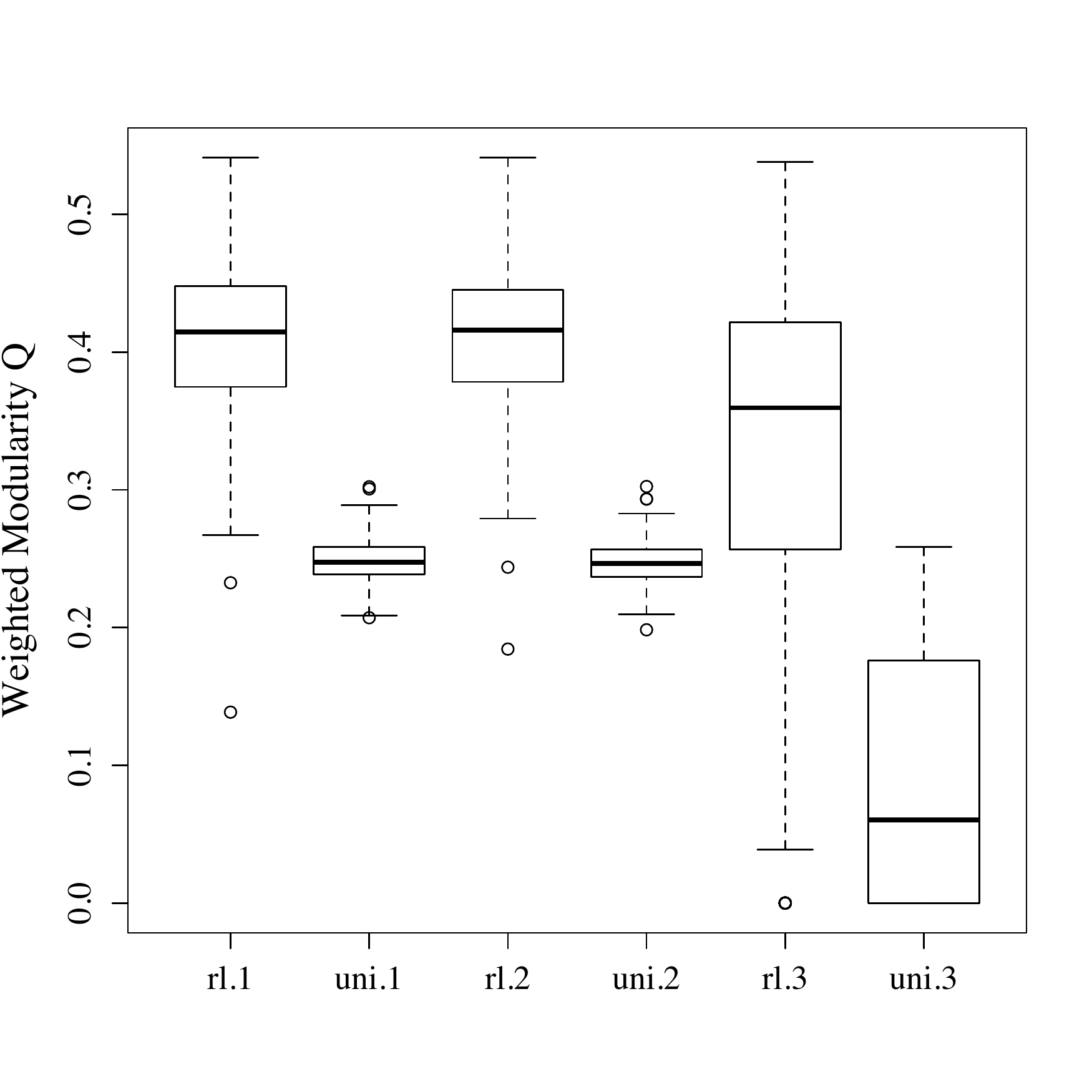}
  \vspace{-0.3cm}
 \caption{Weighted modularity scores $Q$ for the best community clusterings found by the greedy optimization algorithm (left), by the spin-glass ground-state algorithm (center), and by the MCL algorithm (right). The first two are applied to the filtered and undirected but weighted graphs, whereas MCL applies to the original unfiltered weighted and directed ones.}
\label{MCLvsGreedy}
\end{center}
\end{figure}

To end this section, we now briefly present the results obtained through the use of the MCL algorithm, as explained in sect.~\ref{net-analys}. In Fig.~\ref{MCLvsGreedy} we report
the modularity scores for the two algorithms we used on the weighted undirected filtered networks (greedy modularity optimization and spin-glass model), and for MCL on the complete weighted and directed LONs.

To be able to compare modularity scores in a consistent way, we took each algorithm's best found community subdivision and we computed the weighted variant of its $Q$ value on the original unfiltered networks. From the figure,
we observe that the results are qualitatively the same, i.e. real-like instances give rise to optima networks with  a more modular structure than uniform ones, even if
MCL's results have higher variance and for uniform instances sometimes no community is found at all. 
Indeed, MCL faces very dense networks and, to get useful results, the algorithm rescales link weight values in a preprocessing phase.
In the end, we feel that both methods give comparable results but filtering out the networks and then searching for clusters gives 
rise to more stable partitions.

\subsection{Discussion}

\begin{figure}[ht!]
\begin{center}
 \includegraphics[width=0.75\textwidth]{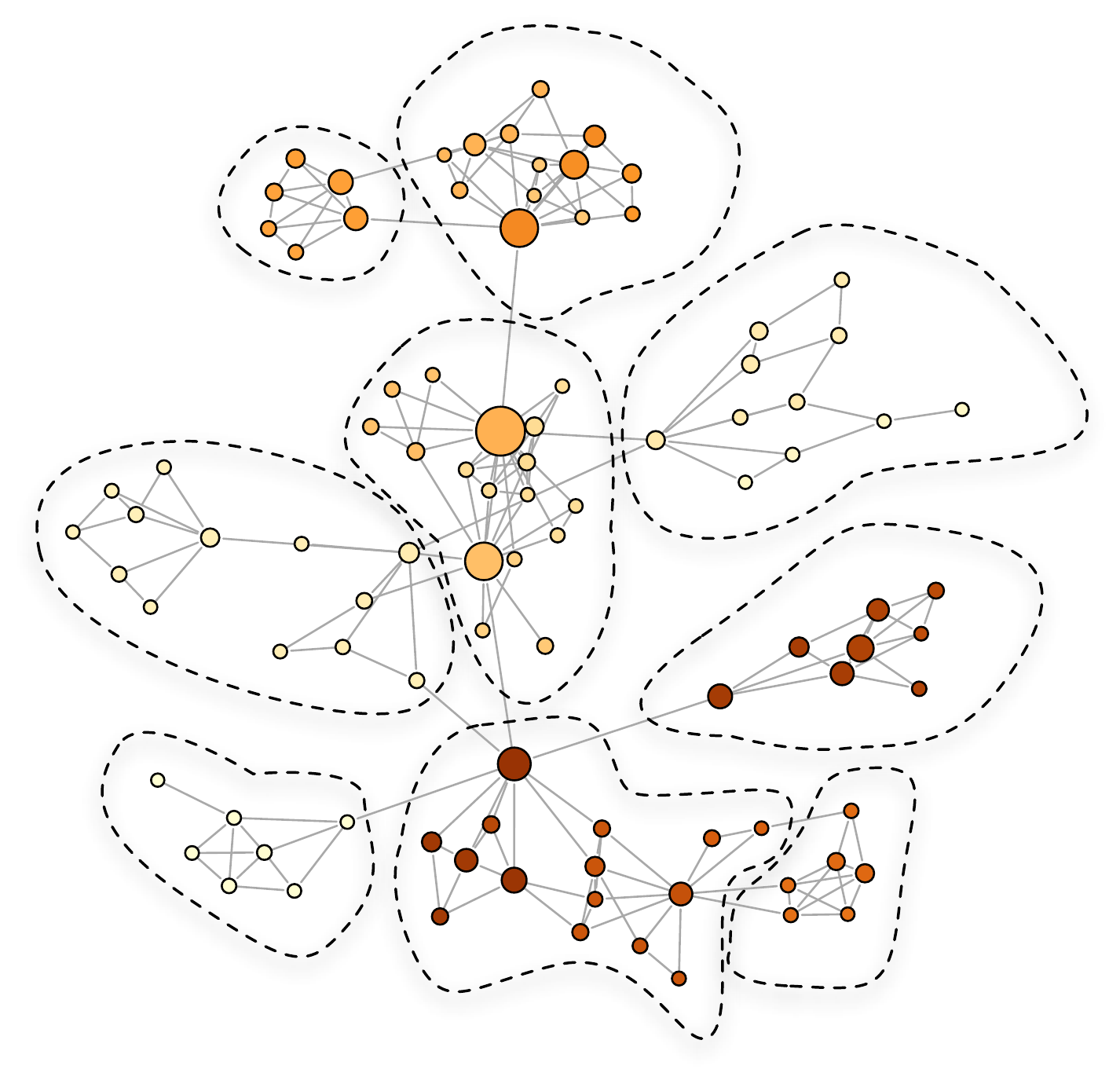}
  \vspace{-0.3cm}
 \caption{Community structure of the LON of a real-like instance. The cluster partition
 found is highlighted. Node sizes are proportional to the corresponding basin size. Darker colors
 mean better fitness (lower).}
\label{comm-rl}
\end{center}
\end{figure}

\begin{figure}[ht!]
\begin{center}
 \includegraphics[width=0.75\textwidth]{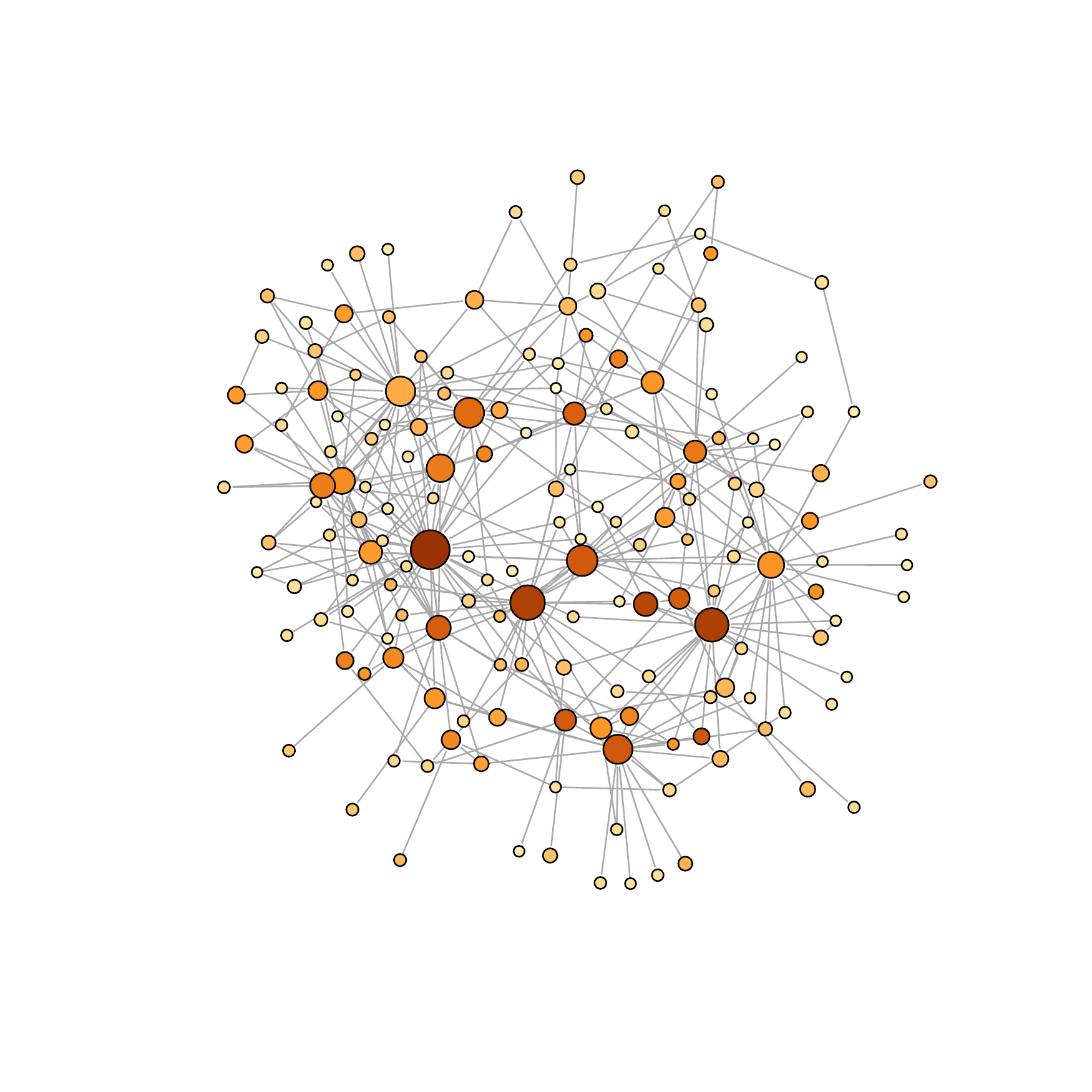}
  \vspace{-0.3cm}
 \caption{Cluster structure of the LON of a random uniform instance. Clusters are less well separated (see text) and cannot
 be clearly highlighted. Node sizes are proportional to the corresponding basin size. Darker colors
 mean better fitness.}
\label{comm-uni}
\end{center}
\end{figure}

From all the previous statistical tests it is apparent that real-like instances have significantly more cluster structure than the class
of random uniform instances of the QAP problem. This can be appreciated visually by looking at Figs.~\ref{comm-rl} and~\ref{comm-uni}
where the community structures of the LON of two particular instances are depicted. Although these are the two particular cases
with the highest $Q$ values of their respective classes,
the trends observed are general.

Fig.~\ref{comm-rl} shows the minima community structure of an instance of the real-like class.
One can see that groups of minima are rather recognizable and form well separated clusters (encircled with dotted lines), which is also
reflected in the high corresponding modularity value $Q=0.79$. Contrastingly, Fig.~\ref{comm-uni} represents a case drawn
from the class of random uniform instances. The network has communities, with a $Q=0.53$, although they are hard to represent graphically, and thus are not shown in the picture. 

In the figures, the diameter of the nodes is drawn proportional to the size of the corresponding basin of attraction in the
fitness landscape. As for the fitness values, the lower (better) the value, the darker the node is. In~\cite{QAP-cec-10} it was found that
there is a positive correlation between fitness values and the corresponding basin size, especially for the random uniform
problem instances. This effect is qualitatively easy to spot on the figures.
The results of this community study, together with~\cite{QAP-cec-10}, sheds light on an open problem in the structure of difficult combinatorial landscapes.
The basin sizes of these problems have been often taken either constant or uniformly distributed at random
for mathematical reasons of simplicity~\cite{garnier01}. However, this is far from being the case for the QAP problem~\cite{QAP-cec-10} and the 
NK landscapes~\cite{kauffman93,pre09} at least. While this conclusion cannot be generalized easily, it could also hold
for other families of difficult combinatorial problems based, as the QAP, on permutation neighborhood such as the Traveling Salesman
Problem (TSP) for example.

From the point of view of the clustering of solutions in the problem fitness landscapes, it has been conjectured that QAP
landscapes fall essentially into two classes: non-structured, with local optima randomly scattered through the search space,
and ``massif central'', with the optima clustered into few localized regions~\cite{merz04}. The suggestion was based on
sampling the fitness landscapes with random walks and measuring entropy and mean distances among local optima.
Our community analysis  supports and goes beyond this intuition providing a richer and fitness-independent global 
view of the full mesoscopic structure of the solutions network. 

\begin{figure}[h!]
\begin{center}
 \includegraphics[width=0.49\textwidth]{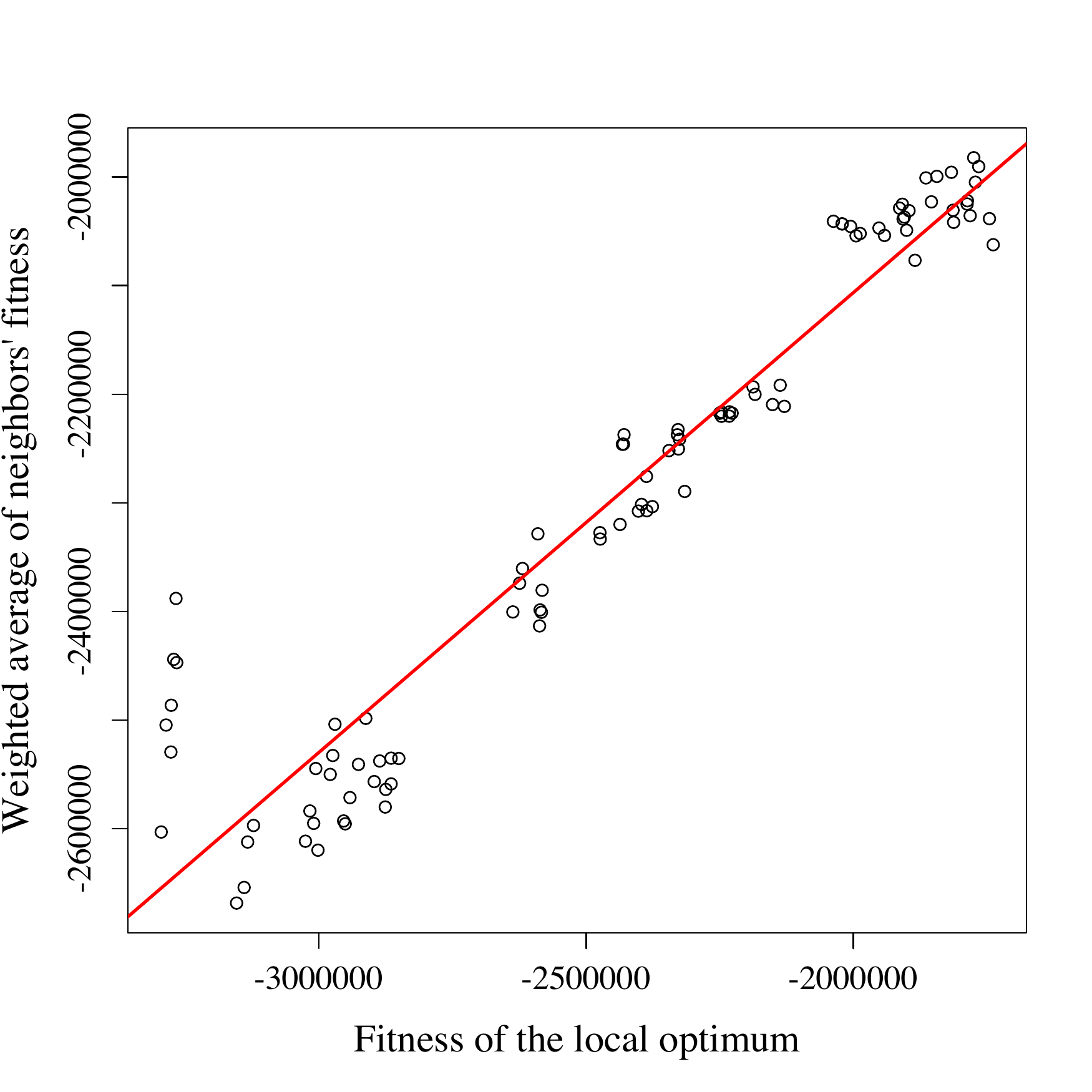}
 \includegraphics[width=0.49\textwidth]{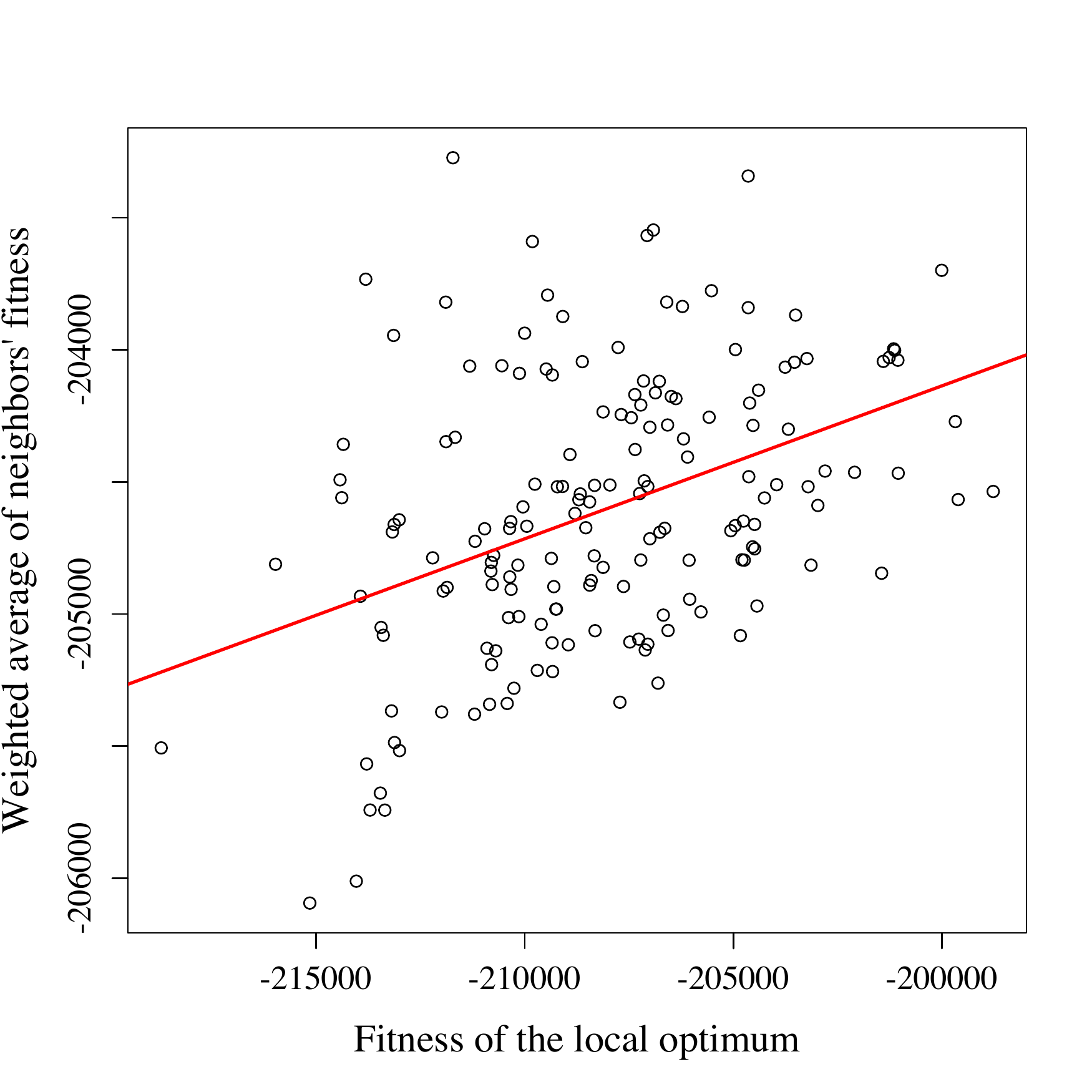}
  \vspace{-0.3cm}
 \caption{Scatterplots and regression lines of the mean fitness of the nearest neighbors of a vertex having fitness $f$. 
 Left image: real-like instance corresponding to
 Fig.~\ref{comm-rl}. Right image: uniform instance corresponding to Fig.~\ref{comm-uni}. Mean neighbor fitness
 is weighted with the transition probabilities of the corresponding outgoing links.  }
\label{correlation}
\end{center}
\end{figure}

To complement the previous purely topological view,
it is useful to investigate the correlation of fitness values between neighboring vertices in the unfiltered LONs. Figure~\ref{correlation} shows
scatterplots of the correlation between the fitness of a given node and the average fitness of its first neighbors in the graph. The plots
correspond to the two particular cases shown in Figs.~\ref{comm-rl} and Fig.~\ref{comm-uni}.
It is apparent that the network is definitely assortative with respect to fitness in the real-like case and slightly so in the uniform case.
Indeed, regression lines show that the correlation is positive in both cases but it is higher in the real-like one, which also has a lower variance.
Although we show results for two particular networks here, we have computed averages over all networks and the trend is the same:
in the unfiltered case we get a Spearman correlation coefficient $r=0.7677$ for the real-like instance class and $r=0.3992$ for the uniform one.
The regression-line slopes are, respectively, $0.249$ and only $0.055$. 
Interestingly, for the filtered backbone networks, fitness becomes disassortative for uniform instances ($r=-0.3459$) whereas it is
still assortative for the real-like case ($r=0.2286$) .

Although it is outside the scope of the present study,
we should like to mention that the above results may have deep consequences on the heuristics used to search combinatorial
spaces such as those described here. For example, in the case of random uniform instances, the landscape has little
structure with a small range of fitness values, so that a standard local search heuristic such as best improvement or
first improvement will quickly find
a satisfactory solution. In contrast, in the landscapes generated by real-like instances, there will be almost separated
groups of minima and thus a parallel-search heuristic that simultaneously explores several regions of the search
space, or a restart strategy like iterated local search, could be more effective. Also, in this case the use of a non-local move operator would perhaps be beneficial.

\section{Summary and Conclusions}
\label{concl}

In this work, we have presented a new methodology to study the structure of the configuration spaces of hard combinatorial
problems. It essentially consists in building the network which has as nodes the locally optimal configurations and
as edges the weighted oriented transitions between optima.
In particular, here we applied the approach to the detection of communities in the optima networks produced
by two different classes of instances of the QAP, which is a hard combinatorial optimization problem.
We provided evidence for the fact that the two problem instance classes give rise to very different configuration spaces
and thus their optima networks are also distinct. 
These results are consistent for both the filtered and unfiltered networks analyzed.
For the so-called real-like class of instances the networks possess
a clear modular structure, while the optima networks belonging to the class of random uniform instances are less well 
partitionable into clusters. This has been convincingly supported by using several statistical tests.

The purely topological view of the distribution of local optima in
the two problem instances classes, was complemented by a study
indicating a higher correlation between the fitness values of
neighboring vertices in the real-like case. The consequences for heuristically searching the corresponding problem 
spaces are at least twofold. First, in the case of random
uniform configuration spaces a simple local heuristic search, such as hill-climbing, should be sufficient to quickly
find satisfactory solutions since they are homogeneously distributed. In contrast, in the real-like case they are much
more clustered in regions of the search space. 
This leads to more modular optima networks and using multiple parallel searches
would probably be a good strategy. These ideas clearly deserve further investigation. Also,
in this paper we have used exhaustive search of the configuration spaces in
order to build the LONs. This is adequate but it can be done only for relatively small instances, as the space
size increases super-exponentially. Thus, our next step will be to develop efficient sampling techniques for larger
problem sizes.

\paragraph*{Acknowledgments}
Fabio Daolio and Marco Tomassini
gratefully acknowledge the Swiss National Science Foundation for financial support under grant number 200021-124578. F.D. also thanks prof. J\'er\^ome Goudet for his valuable suggestions on statistical tests.

\noindent Gabriela Ochoa gratefully acknowledges the British Engineering and Physical Sciences Research Council (EPSRC) for financial support under grant number EP/D061571/1.


\end{document}